\documentclass[letterpaper, 10 pt, conference]{ieeeconf}  

\IEEEoverridecommandlockouts                              

\usepackage{graphicx}
\usepackage{color}
\usepackage{subfigure}
\usepackage{url}
\usepackage{amsmath}
\usepackage{comment}
\usepackage{cite}
 
\title{\LARGE \bf A light-weight, multi-axis compliant tensegrity joint}

\author{Steven Lessard$^{1,*}$, Jonathan Bruce$^{1,*}$, Erik Jung$^{1,*}$, Mircea Teodorescu$^1$, Vytas SunSpiral$^{2,*}$, and Adrian Agogino$^{1, *}$
\thanks{*Authors with the NASA Ames Dynamic Tensegrity Robotics Lab, 
        Moffett Field, CA 94035}
\thanks{$^{1}$University of California, Santa Cruz,
        Santa Cruz, CA 95064, USA
        {\tt\small slessard@ucsc.edu, jbruce@soe.ucsc.edu, mteodore@ucsc.edu}}%
\thanks{$^{2}$Stinger Ghaffarian Technologies,
        Greenbelt, MD 20770, USA
        {\tt\small vytas.sunspiral@nasa.gov}}%
}

\begin{document}

\maketitle
\thispagestyle{empty}
\pagestyle{empty}

\begin{abstract}
In this paper, we present a light-weight, multi-axis compliant tenegrity joint that is biologically inspired by the human elbow.
This tensegrity elbow actuates by shortening and lengthening cable in a method inspired by muscular actuation in a person.
Unlike many series elastic actuators, this joint is structurally compliant not just along each axis of rotation, but along other axes as well.
Compliant robotic joints are indispensable in unpredictable environments, including ones where the robot must interface with a person.
The joint also addresses the need for functional redundancy and flexibility, traits which are required for many applications that investigate the use of biologically accurate robotic models.

\end{abstract}

\section{INTRODUCTION}
 
For many applications of robotic arms, flexibility and structural compliance are critical.
In unpredictable environments especially, these two attributes provide robots with the ability to robustly handle stresses.
These features, however, are difficult to express through traditional robotics.
Even typical series elastic actuators comply with impedances only along their axes of rotation.
Some soft robots however, such as those which adhere to the principles of tensegrity (``tensile-integrity"), are typically better able to to flex under stress and absorb impacts from many directions \cite{Motro2009, Fest2004}.
These soft-bodied robots inherently resist impulses better than traditional robotic alternatives because their tension networks distribute applied forces more evenly throughout the structure.
As a result, even lightweight robots can withstand relatively large impacts and loads \cite{Skelton2009}.

These properties of tensegrity arms also address the need for biomimetic robotic arms to passively handle large moments applied out of phase of the main axis of rotation.
Loads carried by arms can have detrimental effects on the structure of the arm if the induced moment is large enough.
When these loads are applied at an unanticipated angle, unprepared systems will fail.
Human arms, despite their inherent structural levers \cite{dern1947forces}, observably avoid this shortcoming by complying with impacts from many different angles
This emerging tensegrity theory on biomechanics explains this phenomenon as the arm flexibly complying with impedances along multiple axes.
The structural compliance observed in tensegrity robots, such as ours, mitigates this danger by mimicking human joints and distributing loads and stresses along multiple axes.
In addition, the parallel nature of many tension elements within the tensegrity system prevent the failure of single components from destroying the entire structure.
  
\begin{figure} 
\centering
{\includegraphics[width=0.9\linewidth]{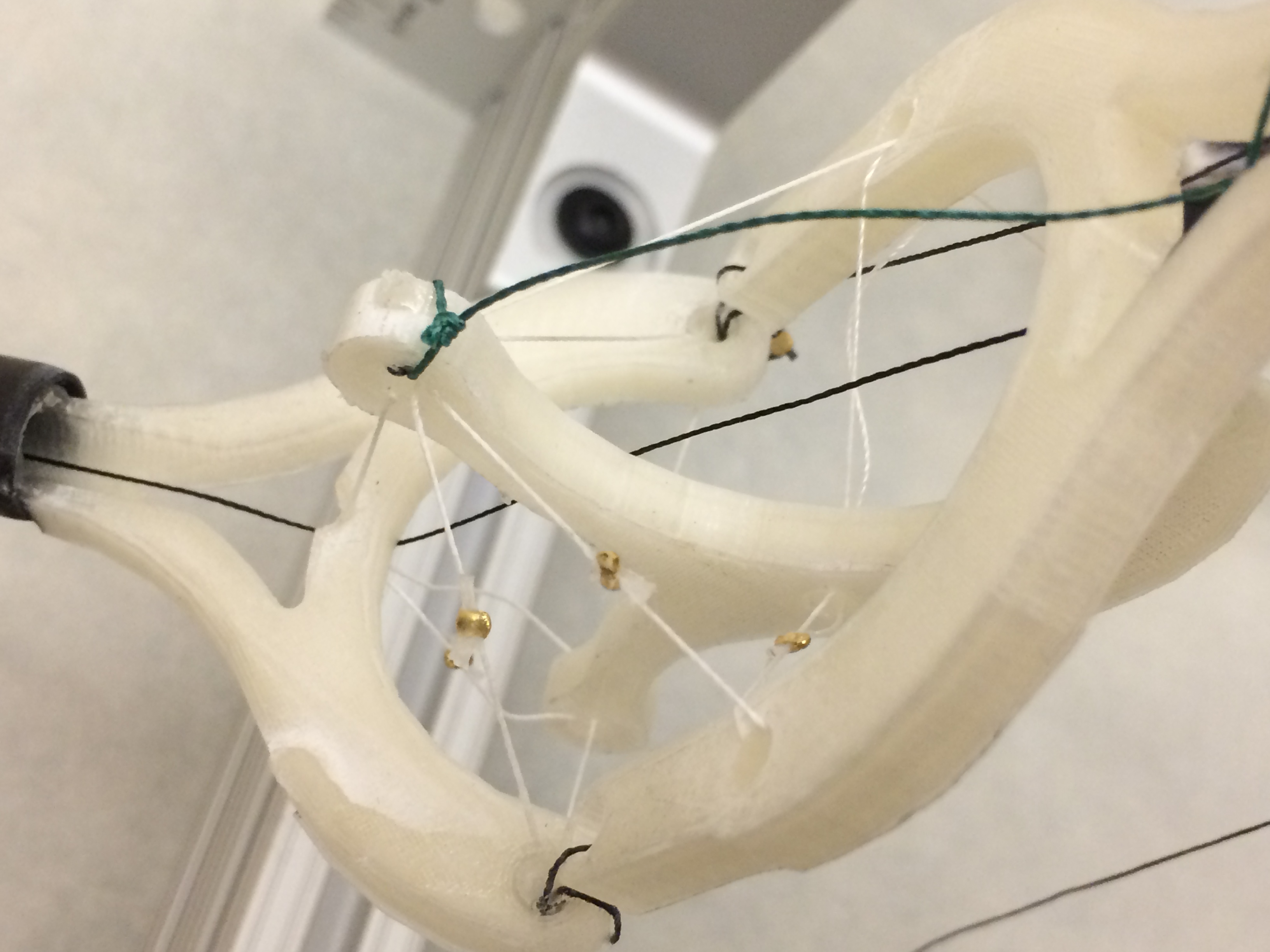}}
\caption{A light-weight, multi-axis compliant tensegrity joint.
This joint can actively control the pitch and the yaw of the arm to which it is attached by contracting and releasing strings which are paired antagonistically.
These active string pairs can also be tuned for stiffness, creating a variable level of flexibility within the joint.}
\label{phys_closeup}
\end{figure}
 
Our multi-axis approach (Figure \ref{phys_closeup}) also has the additional bonus of being able to articulate the arm in a manner unparalleled biologically.
Although anatomically correct human arms generate yaw motion through shoulder muscles, our robotic is able to generate this movement directly from the elbow.
Multi-axis movement in the elbow both protects the joint from imposed moments and allows it a larger range of motion.

The flexibility and structural compliance in our particular arm can be attributed to the tensegrity joint we have developed.
Unlike a simple hinge, this joint features two independent axes.
Actuation along these axes control the pitch and the yaw of the end-effector of the arm.
For each axis, we have a pair of antagonistic cables which actively control motion in opposing directions around a particular axis.
In addition to these actuated cables, there are five additional pairs of antagonistic, passive cables in the joint for stability and force distribution.
The pairing scheme of our tensegrity joint is inspired by the muscular and fascial connections within the human elbow.
Because all actuated movement of the arm is generated by pulling and releasing cables, the motors can be located off of the robot itself.
This lightens the weight of the robotic arm, allowing it to be actuated more easily.

In this paper, we first discuss the background that demands a more flexible and compliant robotic arm joint.
Then, we present the design of our system: both the specific layout of the elbow and the methods to control it.
We then show how we use of the NTRT simulator for rapid design and testing of tensegrity joints.
Next, we illustrate the capabilities of our constructed model as well as the hardware we have used to verify our simulated results.
We conclude with a summary of our contribution as well as the focus of future work.

\section{BACKGROUND INFORMATION}

\subsection{Biological Inspiration}
In human joints, bones, muscles, and fascia connect to form intricate, heterogenous systems.
Each type of tissue is unique in both its structural and material properties.
Because of this diversity, human joints support many functions which range in strength, precision, and support.
These joints are also simultaneously durable and structurally compliant, imporving their ability to react to impedances.
Although relatively little research has been performed regarding fascia as a major component when building human-based robotic sytems \cite{van2009architecture}, its role as a connector between major compression elements in the body (i.e. bones) and major tension elements within the body (i.e. muscles) cannot be overlooked when designing biomimetic joints.
Robots that possess these abilities can likewise accomplish numerous, and often unanticipated, tasks. 

\subsection{Tensegrity Structures and Robotics}
Our robot is based upon the tensegrity design paradigm: a biologically founded method of designing both passive and active structures.
Tensegrity structures are composed of two main components: compression elements and the tension elements that connect them in suspension.
In a traditional tensegrity structure, compression elements are simple rods, whose end points act as hubs through which tension elements, such as cables, pass.
As a result, tensegrity structures invariably feature no two compression elements directly contacting each other.
The resulting structure is flexible and structurally compliant.
Impact forces are distributed throughout the entire structure, decreasing the often detrimental consequences of strong mechanical impulses.
As a result, sudden impedances can be handled relatively elegantly and passively by the structure itself.
In the context of a tensegrity arm, this means that joints can better resist the torques applied by potential levers (such as the component analogous to a forearm).

As research into tensegrity structures has increased, so has the demand for active tensegrities which model specific joints within the human body.
Turvey and Fonseca discuss the need to build and study active tensegrity joints of the elbow based upon the passive elbow design of Scarr \cite{turvey2014medium, scarr2012consideration}.
In this design, the elbow is treated as compression elements (i.e. metal rods) held in equillbirium by string.
The compression elements are meant to emulate bones and the strings to emulate muscles and fascia.
Although this model does not heavily focus on anatomical accuracy, it illustrates that the basic hinge seen in traditional elbow models can be redesigned with the tensegrity principles.

\begin{figure} 
\centering
{\includegraphics[width=0.9\linewidth]{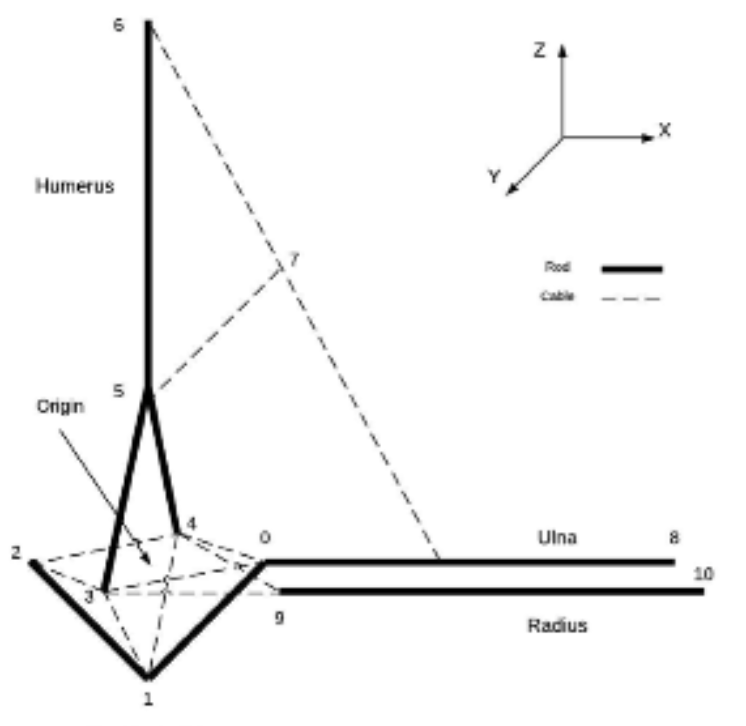}}
\caption{The design for the first passive structure simulated. 
This was based upon Scarr's passive tensegrity elbow\cite{scarr2012consideration}}
\label{originalpassivedesign}
\end{figure} 
 
\subsection{Structural Compliance in Robotic Joints}
Traditional robots are rigid and often cannot bear significant weight when loaded at an unanticipated angle.
As a result, they often lack the ability to resist impedances from unpredictable directions.
This is an especially important problem since many traditional robotic joints have only one axis per joint.
For robots in environments where not every impedance can be predicted, like when imitating human activity, the structure of that robot must be able to endure unanticipated forces.
This handling can be done actively through intelligent control or passively through structural compliance.

Series elastic actuators are one solution to passive impedance resistance.
These actuators are built with an elastic, like a spring, along their axis of actuation.
These springs serve as a cushion which can resist sudden strains, thus preventing system failure.
One downside of these components is that they are only able to resist strains along their axis of actuation.
In systems where forces may not necessarily be applied along this axis, the internal spring does little to mitigate the shock.

Other solutions to the problem of structural compliance lie in the materials used for the actuator.
Pneumatic actuators, like McKibbens artificial muscles, inflate and deflate elastically \cite{tondu2012modelling}.
In this regard, they function very similarly to the skeletal muscle found in many animals.
Some soft robots are constructed using soft materials, such as dielectric elastomers.
These actuators distribute stresses throughout their structure well, but are limited in their flexibility. 

While these solutions seek to improve the compliance of actuators, the tensegrity principle offers a solution on the structural level.
They are designed to be passively stable and since all tension elements are connected in a network, each strain on a tension element propagates to the other tension elements.
As a result, robots which adhere to the tensegrity principles can resist impedances in multiple dimensions.

\section{SYSTEM DESIGN AND CONTROL}
To construct the tensegrity structure behind our light-weight, multi-axis joint, we defined the characteristics and placement of the compression elements and the tension elements. 
After multiple iterations of design, we developed the following structure which enables the ability to actuate along multiple axes within a single elbow joint.

\subsection{Compression Elements}
The designs of the three compression elements (made of plastic and carbon filament, in our prototype) in the tensegrity elbow are inspired by human arm bones.
Our elbow, however, segregates compression elements slightly differently than true bones in the human arm.
The first compression element mimics the humerus and is located above the elbow joint itself.
The compression element within the tensegrity joint is analogous to the olecranon (the hook end of the ulna).
The third and final compression element condenses what would be the radius and the ulna into a single piece, forming the "forearm" of the tensegrity arm.
These compression elements define the overall structure of the arm while anchoring and routing the tension elements.

\subsection{Tension Elements}
The tension elements (strings, in our physical prototype) in this design can be segregated into one of two categories: active and passive.

Active strings are coupled into antagonistic pairs, mimicking how true arm muscles are organized.
For every contraction of an active string, its corresponding antagonistic string will relax and lengthen.
                   
The passive strings in the model represent the fasical connections of the elbow: the tendons and the ligaments.
There are five pairs of these passive tension elements, which are arranged to abosorb impact from a large variety of angles.
These tension elements elastically deform according to the actuation of the active muscles and spring the arm back into its original position of equilibrium.
Although these strings are never directly controlled by a motor, they play an integral role in stabilizing the arm as a whole.
The added tensile connections also abosorb shock, preventing the destruction of the active tensile components or even the compression elements.

\subsection{Control}
The Dynamic Tensegrity Robotics Lab at NASA Ames has researched multiple control methods for tensegrities.
Overactuated systems are able to use inverse-kinematics to determine how to actuate the structure \cite{friesen2014ductt}.
In scenarios where systems are underactuated and there exist passive tension elements, machine learning is another viable approach \cite{iscen2015learning}.
Tools, such as neural networks, can make high level decisions based upon an array of inputs (such as sensor readings and pose) to select output motor function for a particular task.
Neurals networks are also powerful tools for this task because they can theoretically be trained ahead of time, decreasing their computational load when the robot operates.

\section{SIMULATION}
\subsection{NASA Tensegrity Robotics Toolkit}
To simulate our tensegrity joint, we used NASA's Tensegrity Robotics Toolkit (NTRT).
NTRT is an open-source simulator for the design and control of tensegrity structures and robots\footnote{Additional information about NTRT can be found at \\ \url{http://irg.arc.nasa.gov/tensegrity/NTRT}}. 
The complex dynamics required to simulate tensegrity structures are calculated using the Bullet Physics Engine (version 2.82).
Real-time video can also be recorded with NTRT.
  
\begin{figure} 
\centering
{\includegraphics[width=0.9\linewidth]{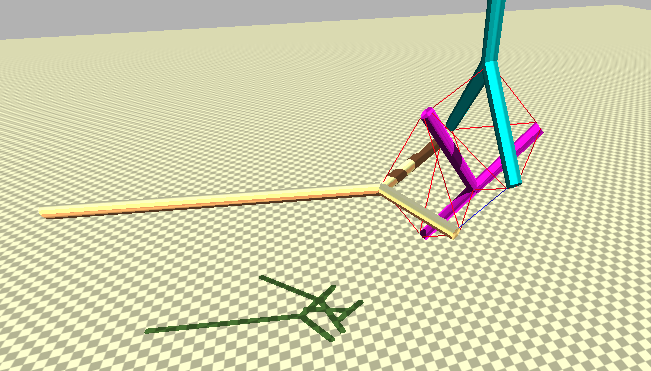}}
\caption{An elbow constructed in NTRT.
The bold colored cylinders are compression elements and the thinner red lines connected to their end points are the cables (tension elements).}
\label{NTRTexample}
\end{figure}
 
In order to create structures within the toolkit, a set of builder tools are utilized which specify geometric rods and connecting cables as a set of Cartesian coordinates. 
These structures can then be specified as substructures and manipulated in three dimensional space as necessary to build complex tensegrity structures (Figure \ref{NTRTexample}).
 
\begin{align}
F = -kX - bV \label{hookeslawdamping}
\end{align}

Within the simulator, cables are modeled as two connected points whose medium lengthens and shortens according to Hooke's Law for linear springs with a linear damping term as well (Equation \eqref{hookeslawdamping}).
Cable control is dictated by functions within a controller class, meaning that the exact length of the cable can be set at each timestep according to a control policy.
Real-world limitations, such as the max acceleration of the motor used and the target velocity of cable lengthening are added to the simulation as well at the structural level.
In addition, maximum and minimum lengths can be applied to each individual cable to prevent unnatural deformations.
These features assert that the robot in simulation is never given extraordinary means to accomplish its goal.
The use of NTRT has already been shown in previous papers to have produced accurate statics and dynamics for SUPERBall, a tensegrity rover designed for extraterrestrial missions \cite{caluwaerts2014design, mirletzcpgs}.
As a result, these simulated robots are able to retain realism.

In addition to modeling the physical aspects of tensegrity structures, NTRT is also useful for testing control policies.
A controller, whether that is a simple, closed-loop periodic function or a more complex machine learning algorithm, can dictate the desired forces in each of the simulated cables, and consequently the desired lengths of each of those cables.
By also implementing restrictions on cable lengths, the full reach of the arm can be tested via simulation.
These simulated models can illustrate the flexibility and compliance in each variation of the robot as they form different poses.
 
\begin{figure} 
\centering
{\includegraphics[width=0.9\linewidth]{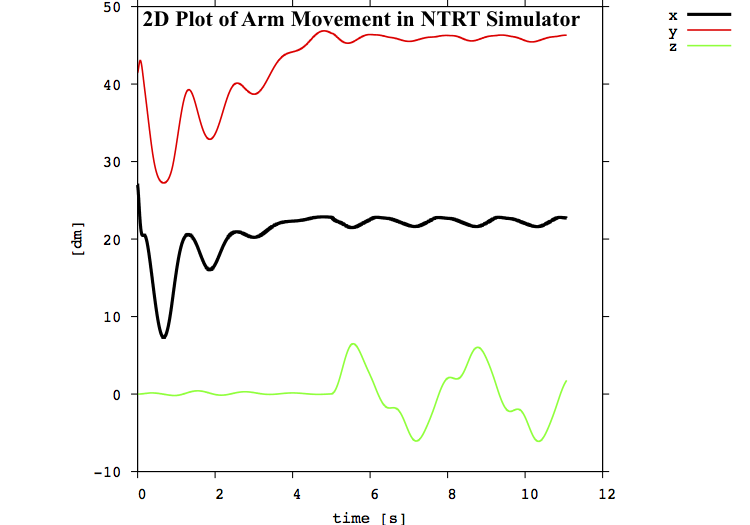}}
{\includegraphics[width=0.9\linewidth]{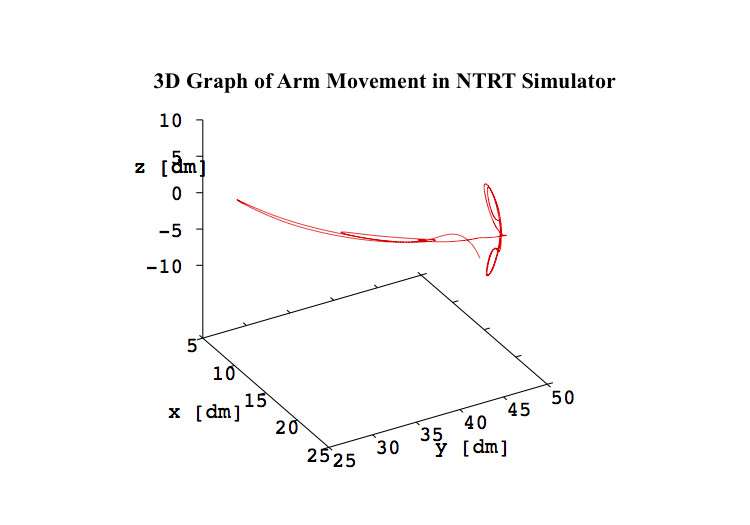}}
\caption{The movement of the end-effector of our simulated tensegrity arm with respect to time.
The plotted path illustrates the arm as it demonstrates pitch motion and then yaw motion.}
\label{plots}
\end{figure}
 
The simulation can also track changes in string lengths similar to how the encoders in the physical prototype measure change in string length.
This common feature means that control policies developed in simulation are more easily portable to the physical prototype.

\section{RESULTS AND DISCUSSION}

\subsection{Actuation Capability}
With our model, we have demonstrated the ability to rotate our arm around two axes independently (Figure \ref{plots}).
By contracting and releasing tension elements, we can change both the pitch and the yaw of our tensegrity arm.

\begin{figure} 
\centering
{\includegraphics[width=0.3\linewidth]{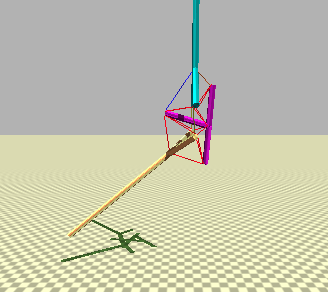}}
{\includegraphics[width=0.3\linewidth]{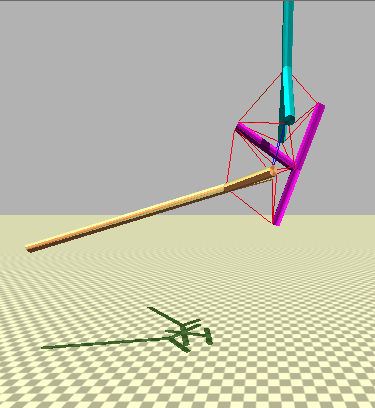}}
{\includegraphics[width=0.3\linewidth]{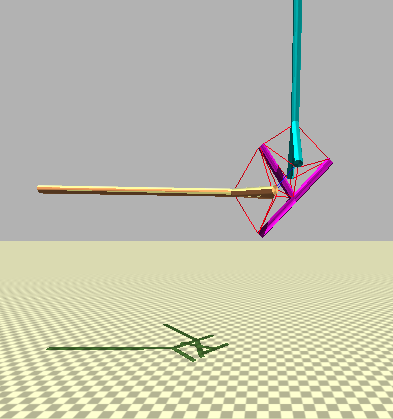}}
\caption{A simulated tensegrity elbow demonstrating pitch movement.
Two cables, mimicking the bicep and tricep in function, act as the antagonistic pair responsible for generating this movement.}
\label{sim_pitch}
\end{figure}

Pitch motion is achieved by changing the lengths of the strings in the antagonistic pair of strings highlighted in Figure \ref{sim_pitch}: one string is shortened while the other is elongated.
The motion shown in this figure mimics bicep contraction and tricep extension.

\begin{figure} 
\centering
{\includegraphics[width=0.3\linewidth]{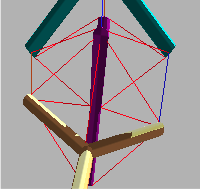}}
{\includegraphics[width=0.3\linewidth]{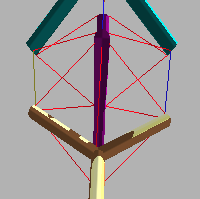}}
{\includegraphics[width=0.3\linewidth]{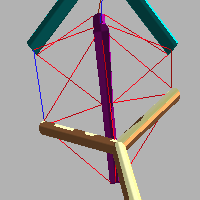}}
\caption{A simulated tensegrity elbow demonstrating yaw movement.
Unlike pitch movement, yaw movement is not generated in the elbow, but instead in the shoulder in an anatomically accurate human arm.}
\label{sim_yaw}
\end{figure}

The yaw of the tensegrity arm is a function of the lengths of the strings in the antagonistic pair of strings highlighted in Figure \ref{sim_yaw}. 
one string is shortened while the other is elongated.
Unlike the demonstrated pitch movement, yaw motion in the elbow does not have a true anatomical corrollary.
Yaw rotation in the human arm is not generated in muscles which connect to the elbow joint, it is instead generated at the shoulder joint.
This gives our tensegrity elbow greater mechanical capability than biological elbows in that aspect.

\subsection{Hardware Validation of Software Models}
After developing the software models of our tensegrity joint, we proceeded to validate these results by constructing physical models.
These models were mostly constructed by 3D-printing compression elements out of polylactic acid (PLA) and by connecting them with either simple string or braided spectra string.
The result was both passive models and an active model, all of which demonstrated the structural robustness expected of tensegrity structures.

\subsubsection{Passive Models}
\begin{figure} 
\centering
{\includegraphics[width=0.9\linewidth]{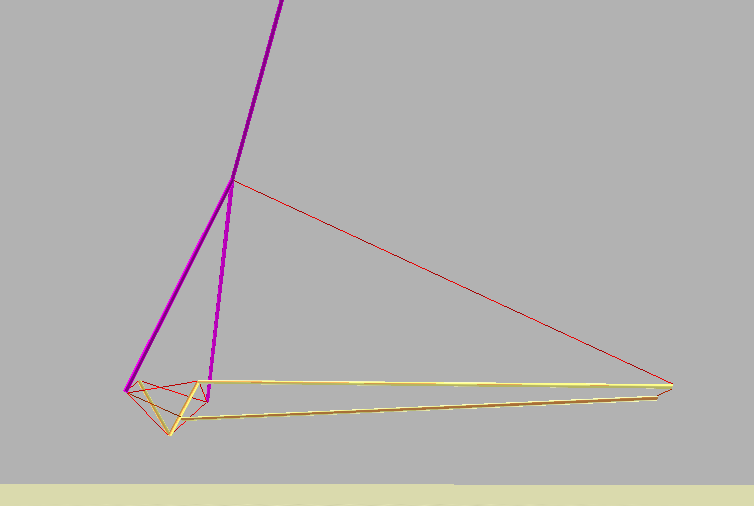}}
\caption{The simulation off of which the initial passive physical prototypes were based.
This was influenced by the designs in Figure \ref{originalpassivedesign}}
\label{simpassive}
\end{figure}

The passive models were first constructed to verify the claims made by Scarr \cite{scarr2012consideration}.
These models did not use motors for actuation, but could still be manually controlled by pulling on specific strings.
Initially, we recreated a simulated version of Scarr's designs in NTRT (Figure \ref{simpassive}) by using the designs from Figure \ref{originalpassivedesign}.
         
\begin{figure} 
\centering
{\includegraphics[width=0.3\linewidth]{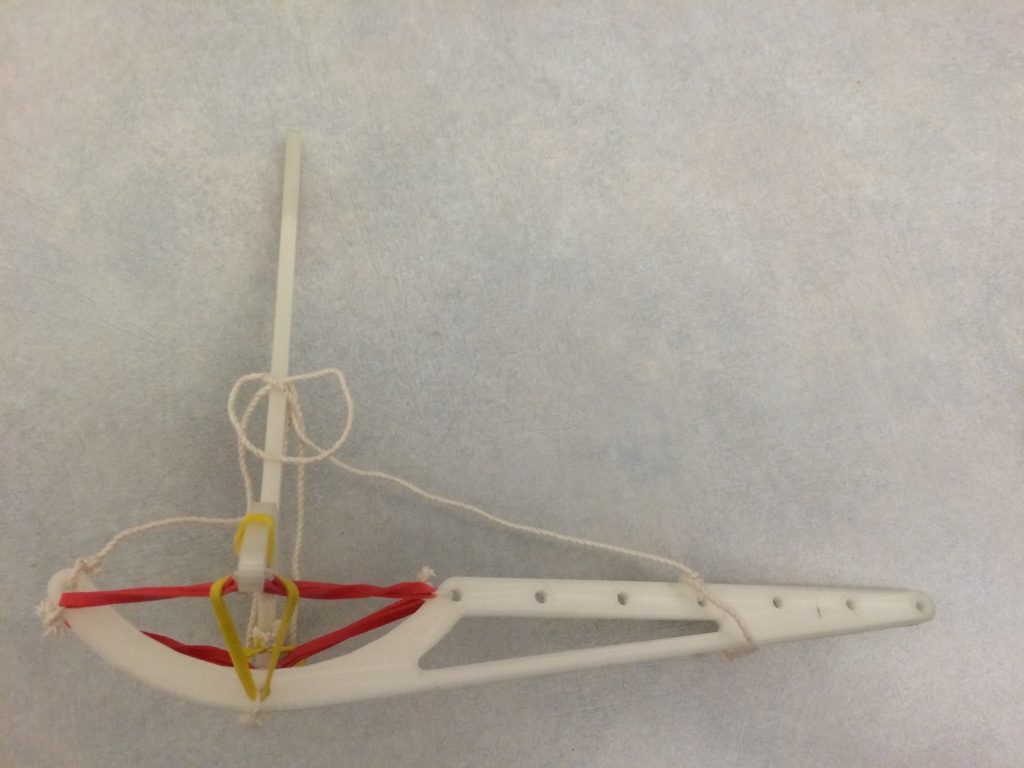}}
{\includegraphics[width=0.3\linewidth]{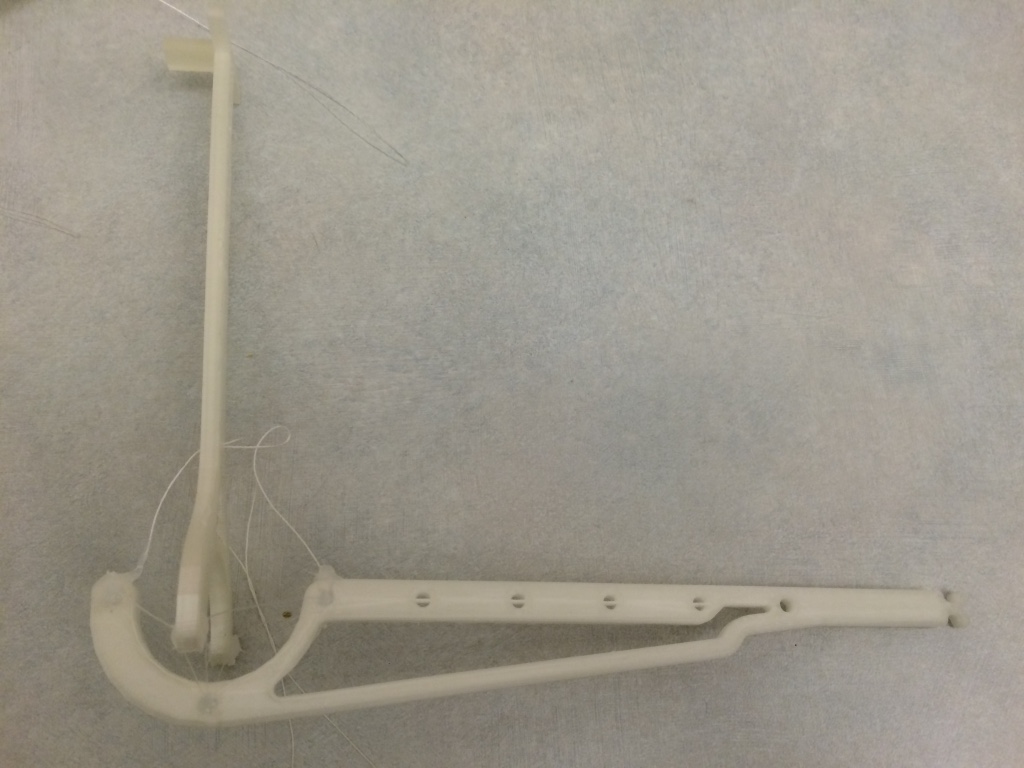}}
{\includegraphics[width=0.3\linewidth]{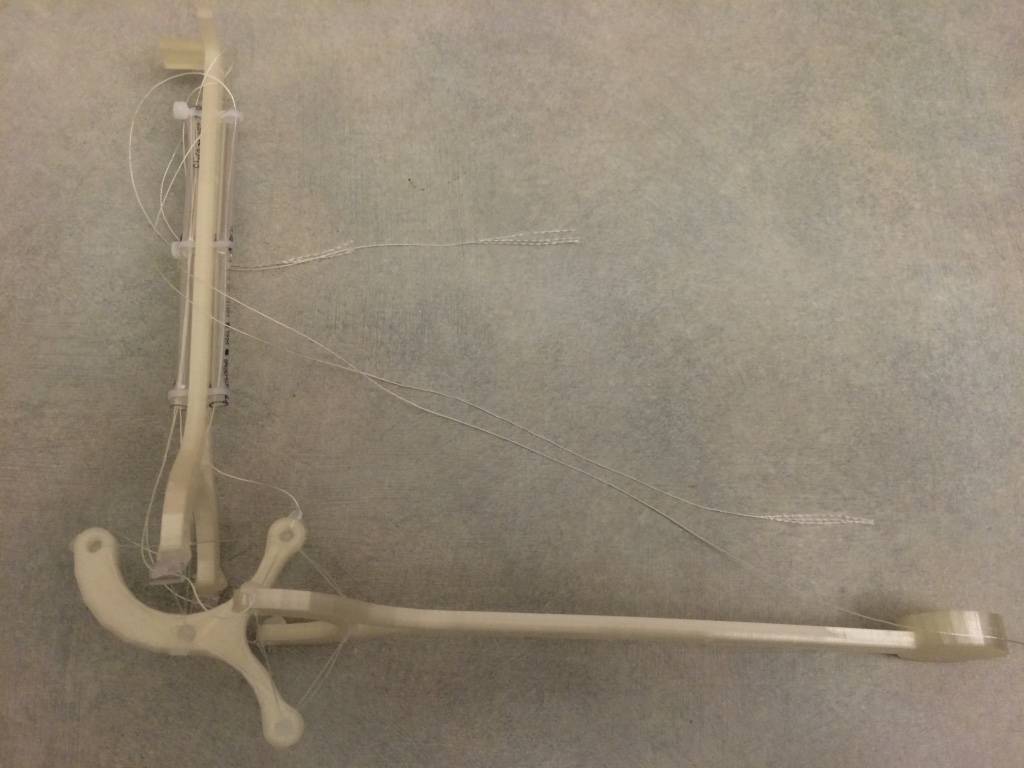}}
\caption{Three iterations of passive tensegrity elbow joints.
These prototypes demonstrate the flexibility and compliance of the tensegrity elbow without using motors.}
\label{passive_prototypes}
\end{figure}

In the first generation of passive models seen in Figure \ref{passive_prototypes}, an elbow joint was constructed using string and elastic bands.
This simple model succesfully demonstrated the ability to move the forearm of our model while still applying external stresses, forcing the model to comply as it moves.

The second model illustrated a miniaturized elbow joint and strings that were routed through the upper compression element.
A basic end-effector was added to this robot that could carry a small payload.

In the third model, we disjointed the end of our forearm (most closely resembly the olecranon in a true elbow).
This new model featured an addition degree of freedom for moving the forearm (yaw).
All active movement in this model can be controlled through 4 strings, all of which are routed through the top compression element.

\subsubsection{Active Model}
\begin{figure} 
\centering
{\includegraphics[width=0.9\linewidth]{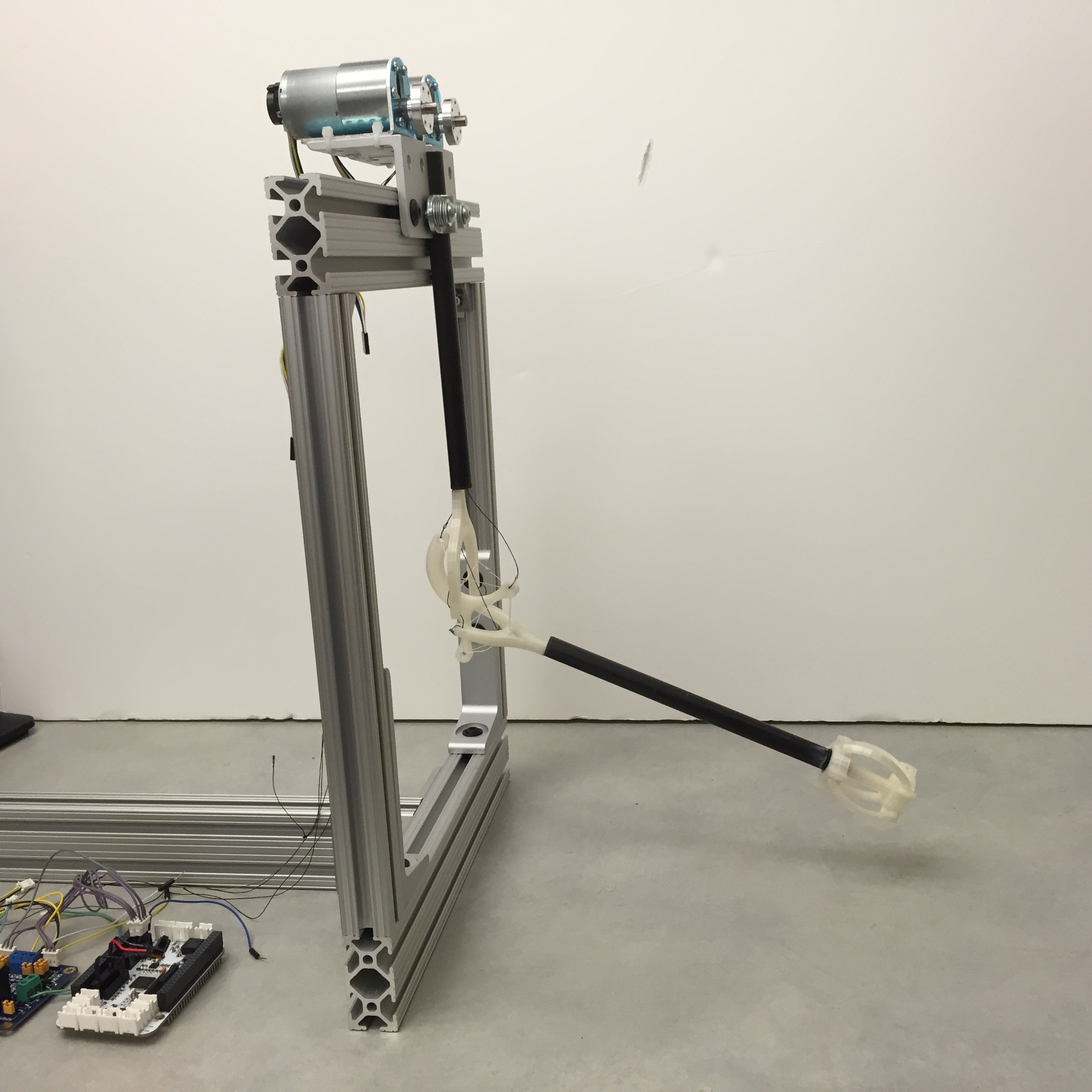}}
\caption{The physical prototype of our tensegrity elbow.
Compression elements are composed of 3D-printed Polylactic Acid (PLA) and tension elements are composed of spectra braided fishing line.
This model is actively controlled by motors seen mounted off of the robot and onto the chassis.}
\label{phys_prototype}
\end{figure}

\begin{table}
\centering
\caption{Weight Constitution of the Active Prototype}
\begin{tabular}{|c|c|c|} \hline
PLA Arm (g)&Both Motors (g)&Combined\\ \hline
$55.6$ & $416.6$ & $472.2$\\ \hline
\end{tabular}
\label{weightconstitution}
\end{table}

The hardware prototype (Figure \ref{phys_prototype}) also reaffirmed the ability of keeping the arm light in weight.
A reduction in weight means that the arm is more reactive to actuation and that it is inherently safer.
For applications that involve or interface with people, safety is an important factor.
The motors were still able to actuate the arm despite being off-loaded onto the chassis.
In this prototype, the motors constituted $88 \%$ of the total mass of the robot (Table \ref{weightconstitution}).
This reduction in weight means that less power is required to operate the tensegrity arm.

\section{CONCLUSION, CONTRIBUTION, AND FUTURE WORK}

By expanding upon the ideas central to the passive tensegrity elbow model constructed by Dr. Graham Scarr, we have developed an accurate software model of a new tensegrity joint.
We have further improved this model by discovering a method in which to construct the tensegrity joint such that is has multiple-axes of actuation.
We have also found a way so that the joint and arm are lighter in weight, thus more reactive to actuation and safer to operate around.
These added features better equip the tensegrity elbow joint for use as an articulation in bio-inspired robotic arms.
Our software model also provides a manner in which to test both the mechanical properties of our robotic arm as well as the efficacy of different control policies.
This simulator produces real-time video of tensegrity joints functioning according to the control policies they have been given.
We have validated these simulations by constructing physical prototypes, both passive and active.
We have also begun to experiment with complex controllers, such as ones governed by neural networks, to tackle the difficult problem of efficiently and effectively actuating an arm with precision and efficiency.

\subsection{Applications in Physical Therapy and Wearable Robotics}
The flexibility and structural compliance seen in these robots are valuable traits when constructing human-oriented devices.
For the field of upper-limb physical therapy and prosthetics, compliant and durable systems with tune-able strength and support are essential.
Many people who own body-powered or electric-powered prosthetics report that their current devices lack the necessary mobility and dexterity required to perform basic tasks \cite{atkins1996epidemiologic}.
Users with body-powered prostheses in this study also cited poor cabling as a major concern.
Emphasizing robust joints in future wearable robotics can potentially address this complaint among upper-limb prosthetic users.
Robust joints will allow prothetics to handle unexpected impedances and higher joint capability, properties which tensegrity joints have illustrated.

In addition, these improvements could be applied to rehabilitative technology as well as prosthetic limbs.
By better understanding the underlying anatomy and its dynamics, wearable devices, such as exoskeletons, can better harmonize with users as they perform bilateral stroke rehabilitation \cite{rosen2007upper}.
As a result, our light-weight, multi-axis joint could be an excellent candidate for future study of wearable robotics as they apply to stroke rehabilitation.
These discoveries may show the potential for our tensegrity elbow to model biological joints in a new manner.

\subsection{Future Work}
Our future work will explore how to more effectively control these tensegrity joints and others through machine learning.
Although simple control is currently possible, high level control decisions, such as moving an end-effector to a position in space while minimizing wobble within the tensegrity structure remains an open question.
High level control could enable the tensegrity elbow to precisely and efficiently manipulate the arm into complex poses by using a generalized algorithm.

In addition to improved control, we will also investigate the possibility of combining this tensegrity joint with other joints in order to construct a more complete and biologically accurate arm.
Finding a better way to model the shoulder as a tensegrity, for example, could significantly improve the functionality of our current arm without sacrificing any of the capability of this elbow joint.




\bibliographystyle{IEEEtran}
\bibliography{tensegrity}

\begin{thebibliography}{10}
\providecommand{\url}[1]{#1}
\csname url@samestyle\endcsname
\providecommand{\newblock}{\relax}
\providecommand{\bibinfo}[2]{#2}
\providecommand{\BIBentrySTDinterwordspacing}{\spaceskip=0pt\relax}
\providecommand{\BIBentryALTinterwordstretchfactor}{4}
\providecommand{\BIBentryALTinterwordspacing}{\spaceskip=\fontdimen2\font plus
\BIBentryALTinterwordstretchfactor\fontdimen3\font minus
  \fontdimen4\font\relax}
\providecommand{\BIBforeignlanguage}[2]{{%
\expandafter\ifx\csname l@#1\endcsname\relax
\typeout{** WARNING: IEEEtran.bst: No hyphenation pattern has been}%
\typeout{** loaded for the language `#1'. Using the pattern for}%
\typeout{** the default language instead.}%
\else
\language=\csname l@#1\endcsname
\fi
#2}}
\providecommand{\BIBdecl}{\relax}
\BIBdecl

\bibitem{Motro2009}
\BIBentryALTinterwordspacing
R.~Motro, ``{Structural morphology of tensegrity systems},'' \emph{ASIAN
  JOURNAL OF CIVIL ENGINEERING BUILDING AND HOUSING}, vol.~10, no.~1, pp.
  1--19, 2009. [Online]. Available:
  \url{http://www.sid.ir/En/VEWSSID/J\_pdf/103820090102.pdf}
\BIBentrySTDinterwordspacing

\bibitem{Fest2004}
E.~Fest, K.~Shea, I.~F.~C. Smith, and M.~Asce, ``{Active Tensegrity
  Structure},'' \emph{Journal of Structural Engineering}, vol. 130, no.~10, pp.
  1454--1465, 2004.

\bibitem{Skelton2009}
R.~E. Skelton and M.~C. De~Oliveira, \emph{Tensegrity Systems},
  2009th~ed.\hskip 1em plus 0.5em minus 0.4em\relax Springer, Jun. 2009.

\bibitem{dern1947forces}
R.~Dern, J.~M. Levene, and H.~Blair, ``Forces exerted at different velocities
  in human arm movements,'' \emph{American Journal of Physiology--Legacy
  Content}, vol. 151, no.~2, pp. 415--437, 1947.

\bibitem{van2009architecture}
J.~van~der Wal, ``The architecture of the connective tissue in the
  musculoskeletal system—an often overlooked functional parameter as to
  proprioception in the locomotor apparatus,'' \emph{International journal of
  therapeutic massage \& bodywork}, vol.~2, no.~4, p.~9, 2009.

\bibitem{turvey2014medium}
M.~T. Turvey and S.~T. Fonseca, ``The medium of haptic perception: A tensegrity
  hypothesis,'' \emph{Journal of motor behavior}, vol.~46, no.~3, pp. 143--187,
  2014.

\bibitem{scarr2012consideration}
G.~Scarr, ``A consideration of the elbow as a tensegrity structure,''
  \emph{International Journal of Osteopathic Medicine}, vol.~15, no.~2, pp.
  53--65, 2012.

\bibitem{tondu2012modelling}
B.~Tondu, ``Modelling of the mckibben artificial muscle: A review,''
  \emph{Journal of Intelligent Material Systems and Structures}, vol.~23,
  no.~3, pp. 225--253, 2012.

\bibitem{friesen2014ductt}
J.~Friesen, A.~Pogue, T.~Bewley, M.~de~Oliveira, R.~Skelton, and V.~Sunspiral,
  ``Ductt: a tensegrity robot for exploring duct systems,'' in \emph{Robotics
  and Automation (ICRA), 2014 IEEE International Conference on}.\hskip 1em plus
  0.5em minus 0.4em\relax IEEE, 2014, pp. 4222--4228.

\bibitem{iscen2015learning}
A.~Iscen, K.~Caluwaerts, J.~Bruce, A.~Agogino, V.~SunSpiral, and K.~Tumer,
  ``Learning tensegrity locomotion using open-loop control signals and
  coevolutionary algorithms,'' \emph{Artificial life}, 2015.

\bibitem{caluwaerts2014design}
K.~Caluwaerts, J.~Despraz, A.~I{\c{s}}{\c{c}}en, A.~P. Sabelhaus, J.~Bruce,
  B.~Schrauwen, and V.~SunSpiral, ``Design and control of compliant tensegrity
  robots through simulation and hardware validation,'' \emph{Journal of The
  Royal Society Interface}, vol.~11, no.~98, p. 20140520, 2014.

\bibitem{mirletzcpgs}
B.~T. Mirletz, R.~D. Quinn, and V.~SunSpiral, ``Cpgs for adaptive control of
  spine-like tensegrity structures.''

\bibitem{atkins1996epidemiologic}
D.~J. Atkins, D.~C. Heard, and W.~H. Donovan, ``Epidemiologic overview of
  individuals with upper-limb loss and their reported research priorities.''
  \emph{JPO: Journal of Prosthetics and Orthotics}, vol.~8, no.~1, pp. 2--11,
  1996.

\bibitem{rosen2007upper}
J.~Rosen and J.~C. Perry, ``Upper limb powered exoskeleton,''
  \emph{International Journal of Humanoid Robotics}, vol.~4, no.~03, pp.
  529--548, 2007.

\end{thebibliography}

\end{document}